\documentclass{llncs}
\usepackage[font=scriptsize]{caption}

\usepackage{times}
\usepackage{helvet}
\usepackage{courier}
\usepackage{amsmath}
\usepackage{breqn}
\usepackage{amssymb}
\usepackage{dsfont}

\usepackage{amsthm}
\usepackage{alltt}
\usepackage{epsfig}
\usepackage{epstopdf}
\usepackage{caption}
\usepackage{wrapfig}
\usepackage{verbatim}

\newcommand{\N}{\mathds{N}}

\title{Encoding Petri Nets in Answer Set Programming for Simulation Based Reasoning}

\author{Saadat Anwar\inst{1}, Chitta Baral\inst{1}\\
	Katsumi Inoue\inst{2}\\
}

\institute{
SCIDSE,  Arizona State University, 699 S Mill Ave, Tempe, AZ 85281, USA \and
Principles of Informatics Research Division, National Institute of Informatics, Japan
}

\begin{document}
\maketitle
\begin{abstract}
One of our long term research goals is to develop systems to answer realistic questions (e.g., some mentioned in textbooks) about biological pathways that a biologist may ask.  To answer such questions we need formalisms that can model pathways, simulate their execution, model intervention to  those pathways, and compare simulations under different circumstances. We found Petri Nets to be the starting point of a suitable formalism for the modeling and simulation needs. However, we need to make extensions to the Petri Net model and also reason with multiple simulation runs and parallel state evolutions. Towards that end Answer Set Programming (ASP) implementation of Petri Nets would allow us to do both.  In this paper we show how ASP can be used to encode basic Petri Nets in an intuitive manner. We then show how we can modify this encoding to model several Petri Net extensions by making small changes. We then highlight some of the reasoning capabilities that we will use to accomplish our ultimate research goal.

\vspace{5pt}
\end{abstract}

\section{Introduction}

The main motivation behind this paper
is to model biological pathways and answer questions of the kind that a biologist would ask. Examples of such questions include ones used in biology text books  to test the understanding of students. We found Petri Nets~\cite{carl1962petri} to be the most suitable starting point for our needs as its graphical representation and semantics closely matches biological pathway diagrams.\footnote{In addition to modeling of biological systems, Petri Nets are also used over a wide range of domains, such as, workflows, embedded systems and industrial control~\cite{van1994modelling,Heiner2006,cortes1999petri,334574,aalst1998workflow,reddy1993petri}.}
However, answering those type of questions requires certain extensions to the Petri Net model and reasoning with multiple simulations and parallel evolutions. Although numerous Petri Net modeling, simulation and analysis systems exist~\cite{jensen2007coloured,SnoopyPN,kounev2006qpme,berthomieu2004tool,nagasaki2010cell,kummer1999renew}, we did not find these systems to be a good match for all our needs. Some had limited adaptability outside their primary application domain, while others, though quite capable, did not offer easy extensibility. Most systems did not explore all possible state evolutions nor allowed different firing semantics. 

To address the twin needs of easy extensibility and reasoning over multiple evolutions we propose encoding Petri Nets using Answer Set Programming (ASP), which is a declarative programming language with competitive solvers~\footnote{See \cite{Clingo,lin2004assat,gebser2007first}} and has been used in modeling domains such as spacecrafts~\cite{nogueira2001prolog}, work flows~\cite{balduccini2009job}, natural language processing~\cite{baral2004using,erdem2009transforming} and biological systems modeling~\cite{dworschak2008modeling,gebser2010bioasp}. In our quest we found ASP to be preferable to process algebra, temporal logics, and mathematical equations applied to Petri Nets. Some of these techniques, like $\pi$-calculus is cumbersome even for small Nets~\cite{van2005pi}, while others, like mathematical equations, impose restrictions on the classes of Petri Nets they can model~\cite{murata1989petri}.

Petri Net translation to ASP has been studied before~\cite{HeljankoNMR,LogicPetriNets}. However, these implementations are limited to specific classes of Petri Nets and have a different focus. For example, ~\cite{HeljankoNMR} used ASP for the analysis of properties of \textit{1-safe} Petri Nets such as reachability and deadlock detection. \textit{1-safe} Petri Nets are very basic in nature as they can accommodate at most one token in a place and they don't allow source/sink transitions or inhibition arcs. As a result their ASP translation does not require handling of token aggregation, leading to simpler conflict resolution and weight-constraints than the general case. In \cite{LogicPetriNets}, a new class of Petri Nets called Simple Logic Petri Nets (SLPNs) is presented and translated to ASP code. This class of Petri Nets uses simple logic expressions for arc weights and positive ground literals as tokens at places. Their implementation does not carry the notion of conflicting transitions, i.e., a token from a source place can be used in multiple transitions in a single firing step. It also does not follow the standard notion of token aggregation, i.e., if the same token arrives at a place from two different transitions, it looses its individual identity and only gets counted as one. Both of these implementations focus on analyzing properties of specific classes of Petri Nets. We design our implementation to simulate general Petri Nets. To our knowledge this has not been attempted in ASP before. However, some of our encoding scheme is similar to their work. Our current focus in this work is less on performance and more on the ease of encoding, extensibility, and exploring all possible state evolutions.

Thus the main contributions of this paper are: In Section~\ref{sec:enc_basic} we show how ASP allows intuitive declarative encoding of basic Petri Nets resulting in a low specification--imple\-mentation gap. We then show, how Petri Net extensions can be incorporated in our encoding by making small changes. In particular, we explore change of firing semantics (Section~\ref{sec:enc_max}), and allowing reset arcs (Section~\ref{sec:enc_reset}), inhibitor arcs (Section~\ref{sec:enc_inhibit}) and read arcs (Section~\ref{sec:enc_query}). We show how our ASP encoding allows exploring all possible state evolutions in a Petri Net simulation and how to reason with those simulation results. We also show how Petri Nets fit into our ultimate goal of answering questions with respect to biological pathways. While the goal of this paper is not to analyze Petri Net properties of reachability, liveness, and boundedness etc., we briefly mention how these can be analyzed using our encoding. 
We now start with a brief background on ASP and Petri Nets.

\section{Background on ASP and Petri Nets}

{\bf Answer Set Programming (ASP)}  is a declarative logic programming language based on the Stable Model Semantics~\cite{StableModels}. Please refer to the Clingo manual~\cite{Clingo} for details of the ASP syntax used in this paper.

\vspace*{0.1in}

{\bf A Petri Net} is a graph of a finite set of nodes and directed arcs, where nodes are split between places and transitions, and each arc either connects a place to a transition or a transition to a place. Each place has a number of tokens (called the its marking). Collective marking of all places in a Petri Net is called its \textit{marking} (or \textit{state}). Arc labels represent arc weights. When missing, arc-weight is assumed as one, and place marking is assumed as zero.

\begin{figure}[htbp]
\centering
\vspace{-30pt}
\includegraphics[width=9cm]{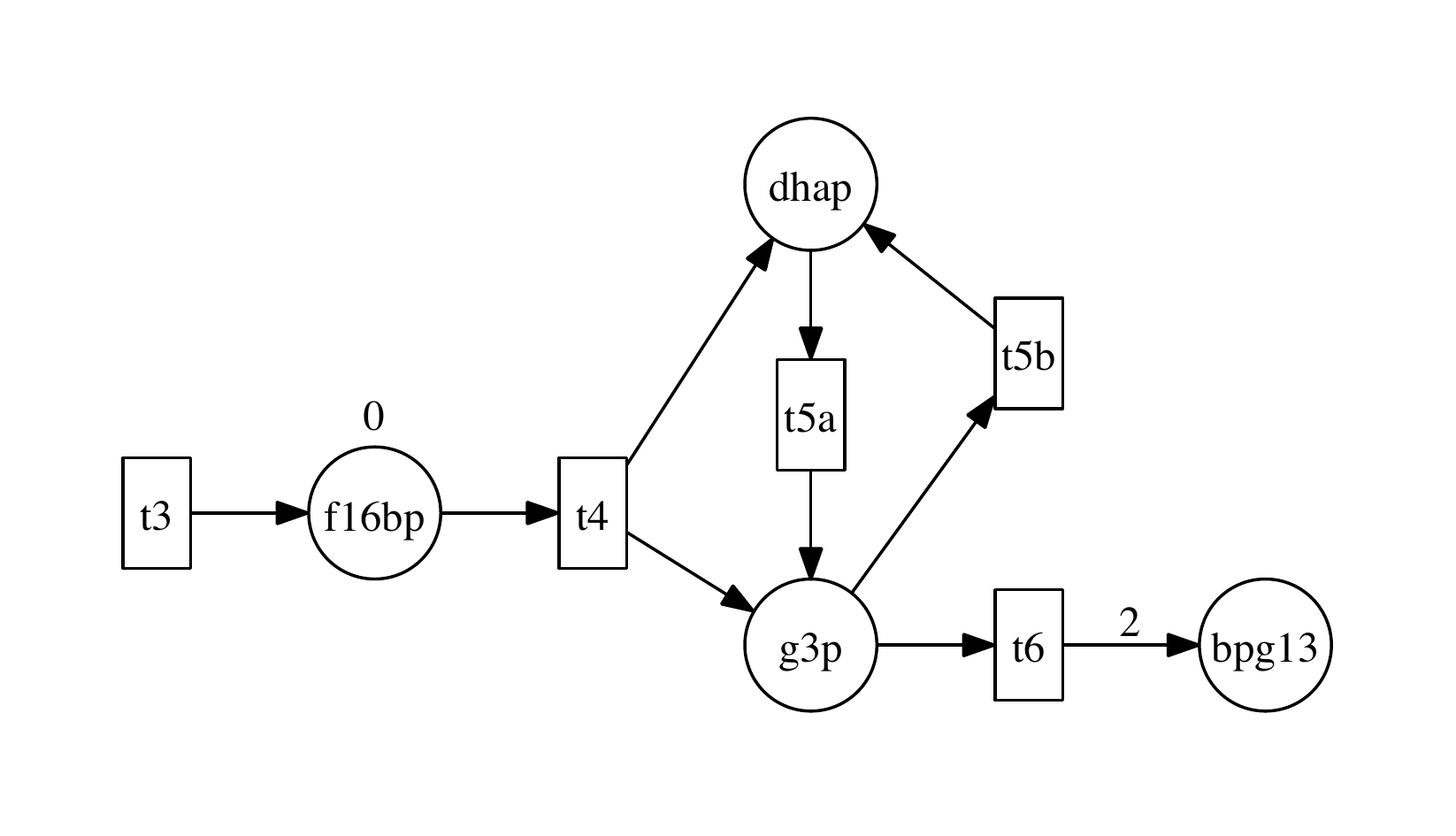}
\caption{Petri Net graph (of sub-section of glycolysis pathway) showing places as circles, transitions as boxes and arcs as directed arrows. Places have token count (or marking) written above them, assumed 0 when missing. Arcs labels represent arc-weights, assumed 1 when missing.}
\label{fig:q1:a}
\end{figure}

The set of place nodes on incoming and outgoing arcs of a transition are called its pre-set (input place set or input-set) and post-set (output place set or output-set), respectively. A transition $t$ is enabled when each of its pre-set place $p$ has at least the number of tokens equal to the arc-weight from $p$ to $t$. An enabled transition may fire, consuming tokens equal to arc-weight from $p$ to $t$ from each pre-set place $p$, producing tokens equal to arc-weight from $t$ to $p$ to each post-set place $p$. 

Multiple transitions may fire as long as they consume no more than the available tokens, with the assumption that tokens cannot be shared. Fig.~\ref{fig:q1:a} shows a portion of the glycolysis pathway~\cite{CampbellBook}, in which places represent reactants and products, transitions represent reactions, and arc weights represent reactant quantity consumed or the product quantity produced by the reaction.

\begin{definition}[Petri Net]
A Petri Net is a tuple $PN=(P,T,E,W)$, where, 
$P=\{p_1, \dots, p_n\}$ is a finite set of places;
$T=\{t_1, \dots, t_m\}$ is a finite set of transitions, $P \cap T = \emptyset$;
$E^+ \subseteq T \times P$ is a set of arcs from transitions to places;
$E^- \subseteq P \times T$ is a set of arcs from places to transitions;
$E= E^+ \cup E^- $; and
$W: E \rightarrow \N \setminus \{0\}$ is the arc-weight function
\end{definition}

\begin{definition}[Marking]\label{def:pn:marking} A marking $M=(M(p_1),\dots,M(p_{n}))$ is the token assignment of each place node $p_i \in P$, where $M(p_i) \in \N$. Initial token assignment $M_0: P \rightarrow \N$ is called the initial marking. Marking at step $k$ is written as $M_k$.
\end{definition}

\begin{definition}[Pre-set \& post-set of a transition]
Pre-set / input-set of a transition $t$ is $\bullet t = \{ p \in P : (p,t) \in E^- \}$, while the post-set / output-set is $t \bullet = \{ p \in P : (t,p) \in E^+ \}$
\end{definition}

\begin{definition}[Enabled Transition]\label{pn:enable}
A transition $t$ is enabled with respect to marking $M$, $enabled_M(t)$, if $\forall p \in \bullet t, W(p,t) \leq M(p)$. An enabled transition may fire. 
\end{definition}

\begin{definition}[Execution]\label{pn:exec}
An execution is the simulation of change of marking from $M_k$ to $M_{k+1}$ due to firing of transition $t$.
$M_{k+1}$ is computed as follows:
\[ \forall p_i \in \bullet t, M_{k+1}(p_i) = M_{k}(p_i) - W(p_i,t) \]
\vspace{-20pt}
\[ \forall p_j \in t \bullet, M_{k+1}(p_j) = M_{k}(p_j)+ W(t,p_j) \]
\end{definition}

\begin{definition}[Conflicting Transitions]\label{pn:conflict}
A set of enabled transitions $T_e = \{ t \in T : enabled_M(t) \}$ conflict if their simultaneous firing will consume more tokens than are available at an input place:
\[
\exists p \in P : M(p) < \displaystyle\sum_{\substack{t \in T_e \wedge(p,t) \in E^-}}{W(p,t)}
\]
\end{definition}

\begin{definition}[Firing Set]\label{pn:firing_set}
A set of simultaneously firing, non-conflicting, enabled transitions $T_k=\{t_{1},\dots,t_{m}\} \subseteq T$ is called a firing set. Its execution w.r.t. marking $M_{k}$ produces new marking $M_{k+1}$ as follows:
\[
\forall p \in P, M_{k+1}(p) = M_k(p) 
- \sum_{\substack{t \in T_k \wedge p \in \bullet t}} W(p,t)
+ \sum_{\substack{t \in T_k \wedge p \in t \bullet}} W(t,p)
\]
\end{definition}

\begin{definition}[Execution Sequence]\label{pn:exec_seq}
An execution sequence is the simulation of a firing sequence $\sigma = T_1,T_2,\dots,T_k$. It is the transitive closure of executions, where subsequent markings become the initial marking for the next firing set. Thus, in the execution sequence $X = M_0, T_0, M_1, T_1, \dots, M_k, T_k, M_{k+1}$, the firing of $T_0$ with respect to marking $M_0$ produces the marking $M_1$ which becomes the initial marking for $T_1$.
\end{definition}
 
\section{Translating Basic Petri Net Into ASP}\label{sec:enc_basic}
In this section we present ASP encoding of simple Petri Nets. We describe, how a given Petri Net $PN$, and an initial marking $M_0$ are encoded into ASP for a simulation length $k$. Following sections will show how Petri Net extensions can be easily added to it.
We represent a {\bf Petri Net} with the following facts:
{\small
\begin{description}
\item[f1:] Facts \texttt{\small place($p_i$).} where $p_i \in P$ is a place. 
\item[f2:] Facts \texttt{\small trans($t_j$).} where $t_j \in T$ is a transition. 
\item[f3:] Facts \texttt{\small ptarc($p_i,t_j,W(p_i,t_j)$).} where $(p_i,t_j) \in E^-$ with weight $W(p_i,t_j)$. %
\item[f4:] Facts \texttt{\small tparc($t_i,p_j,W(t_i,p_j)$).} where $(t_i,p_j) \in E^+$ with weight $W(t_i,p_j)$. %
\end{description}
}

Petri Net {\bf execution simulation} proceeds in discrete time-steps, these time steps are encoded by the following facts:
{\small
\begin{description}
\item[f5:] Facts \texttt{\small time($ts_i$)} where $0 \leq ts_i \leq k$.
\end{description}
}

{\bf initial marking} (or initial state) of the Petri Net is represented by the following facts:
{\small
\begin{description}
\item[i1:] Facts \texttt{\small holds($p_i,M_0(p_i),0$)} for every place $p_i \in P$ with initial marking $M_0(p_i)$.
\end{description}
}

ASP requires all variables in rule bodies be domain restricted. Thus, we add the following facts to capture token quantities produced during the simulation~\footnote{Note that $ntok$ can be arbitrarily chosen to be larger than the maximum expected token quantity produced during the simulation.}:
{\small
\begin{description}
\item[x1:] Facts \texttt{\small num($n$).}, where $0 \leq n \leq ntok$ 
\end{description}
}

A transition $t_i$ is enabled if each of its input places $p_j \in \bullet t_i$ has at least arc-weight $W(p_j, t_i)$ tokens. Conversely, $t_i$ is not enabled if $\exists p_j \in \bullet t_i : M(p_j) < W(p_j,t_i)$, and is only enabled when no such place $p_j$ exists. These are captured in $e1$ and $e2$ respectively:

{
\small
\begin{description}
\item[e1:] \texttt{\small notenabled(T,TS):-ptarc(P,T,N),holds(P,Q,TS),Q<N, place(P), \\trans(T), time(TS),num(N),num(Q).}
\item[e2:] \texttt{\small enabled(T,TS) :- trans(T), time(TS), not notenabled(T, TS).}
\end{description}
}

The rule $e1$ encodes \texttt{\small notenabled(T,TS)} which captures the existence of an input place $P$ of transition $T$ that violates the minimum token requirement $N$ at time-step $TS$. Where, the predicate \texttt{\small holds(P,Q,TS)} encodes the marking $Q$ of place $P$ at $TS$. Rule $e2$ encodes \texttt{\small enabled(T,TS)} which captures that transition $T$ is enabled at $TS$ since there is no input place $P$ of transition $T$ that violates the minimum input token requirement at $TS$. In biological context, $e2$ captures the conditions when a reaction (represented by $T$) is ready to proceed. A subset of enabled transitions may fire simultaneously at a given time-step. This is encoded as:

{\small
\begin{description}
\item[a1:] \texttt{\small  \{fires(T,TS)\} :- enabled(T,TS), trans(T), time(TS). }
\end{description}
}

Rule $a1$ encodes \texttt{\small fires(T,TS)}, which captures the firing of transition $T$ at $TS$. The rule is encoded with a count atom as its head, which makes it a choice rule. This rule either picks the enabled transition $T$ for firing at $TS$ or not, effectively enumerating a subset of enabled transitions to fire. Whether this set can fire or not in an answer set is subject to conflict checking, which is done by rules $a2,a3,a4$ shown later. In biological context, the selected transition-set models simultaneously occurring reactions and the conflict models limited reactant supply that cannot be shared. Such a conflict can lead to multiple choices in parallel reaction evolutions and different outcomes. The next set of rules captures the consumption and production of tokens due to the firing of transitions in a firing-set as well as their aggregate effect, which computes the marking for the next time step:

{\small
\begin{description}
\item[r1:] \texttt{\small add(P,Q,T,TS) :- fires(T,TS), tparc(T,P,Q), time(TS).}
\item[r2:] \texttt{\small del(P,Q,T,TS) :- fires(T,TS), ptarc(P,T,Q), time(TS).} 

\item[r3:] \texttt{\small tot\_incr(P,QQ,TS) :- 
   QQ=\#sum[add(P,Q,T,TS)=Q:num(Q):trans(T)], \\
   time(TS), num(QQ), place(P).}

\item[r4:] \texttt{\small tot\_decr(P,QQ,TS) :- 
   QQ=\#sum[del(P,Q,T,TS)=Q:num(Q):trans(T)], 
   time(TS), num(QQ), place(P).}

\item[r5:] \texttt{\small holds(P,Q,TS+1) :-holds(P,Q1,TS),tot\_incr(P,Q2,TS),time(TS+1),\\
   tot\_decr(P,Q3,TS),Q=Q1+Q2-Q3,place(P),num(Q;Q1;Q2;Q3),time(TS).}
\end{description}
}

Rule $r1$ encodes \texttt{\small add(P,Q,T,TS)} and captures the addition of $Q$ tokens to place $P$ due to firing of transition $T$ at time-step $TS$. Rule $r2$ encodes \texttt{\small del(P,Q,T,TS)} and captures the deletion of $Q$ tokens from place $P$ due to firing of transition $T$ at $TS$. Rules $r3$ and $r4$ aggregate all \texttt{\small add}'s and \texttt{\small del}'s for place $P$ due to $r1$ and $r2$ at time-step $TS$, respectively, by using the \texttt{\small QQ=\#sum[]} construct to sum the $Q$ values into $QQ$.  Rule $r5$ which encodes \texttt{\small holds(P,Q,TS+1)} uses these aggregate adds and removes and updates $P$'s marking for the next time-step $TS+1$. In biological context, these rules capture the effect of a reaction on reactant and product quantities available in the next simulation step. To prevent overconsumption at a place following rules are added:

{\small
\begin{description}
\item[a2:] \texttt{\small consumesmore(P,TS) :- holds(P,Q,TS), tot\_decr(P,Q1,TS), Q1 > Q.} 
\item[a3:] \texttt{\small consumesmore :- consumesmore(P,TS).}
\item[a4:] \texttt{\small :- consumesmore.}
\end{description}
}

Rule $a2$ encodes \texttt{\small consumesmore(P,TS)} which captures overconsumption of tokens at input place $P$ at time $TS$ due to the firing set selected by $a1$. Overconsumption (and hence conflict) occurs when tokens $Q1$ consumed by the firing set are greater than the tokens $Q$ available at $P$. Rule $a3$ generalizes this notion of overconsumption and constraint $a4$ eliminates answers where overconsumption is possible. 

\begin{definition}
Given a Petri Net $PN$ and its encoding $\Pi(PN,M_0,k)$. We say that there is a 1-1 correspondence between the answer sets of $\Pi(PN,M_0,k)$ and the execution sequences of $PN$ iff for each answer set $A$ of $\Pi(PN,M_0,k)$, there is a corresponding execution sequence $X=M_0,T_0,M_1,\dots,M_k,T_k$ of $PN$ such that 
\[ \{ fires(t,j) : t \in T_j, 0 \leq j \leq k \} = \{ fires(t,ts) : fires(t,ts) \in A \} \]
\vspace{-20pt}
\[ \{ holds(p,q,j) : p \in P, q=M_j(p), 0 \leq j \leq k \} = \{ holds(p,q,ts) : holds(p,q,ts) \in A\} \]
\end{definition}

\begin{proposition}\label{prop:basic_enc}
There is a 1-1 correspondence between the answer sets of $\Pi^0(PN,M_0,k)$ and the execution sequences of $PN$.

\end{proposition}

\subsection{An example execution}\label{sec:basic_pn_exec}

Next we look at an example execution of the Petri Net shown in Figure~\ref{fig:q1:a}. The Petri Net and its initial marking are encoded as follows\footnote{\texttt{\{holds(p1,0,0),\dots,holds(pN,0,0)\}}, \texttt{\{num(0),\dots,num(60)\}}, \texttt{\{time(0),\dots,time(5)\}} have been written as \texttt{holds(p1;\dots;pN,0,0)}, \texttt{num(0..60)}, \texttt{time(0..5)},  respectively, to save space.}:
{\small
\begin{verbatim}
num(0..60).time(0..5).place(f16bp;dhap;g3p;bpg13).
trans(t3;t4;t5a;t5b;t6).tparc(t3,f16bp,1).ptarc(f16bp,t4,1).
tparc(t4,dhap,1).tparc(t4,g3p,1).ptarc(dhap,t5a,1).
tparc(t5a,g3p,1).ptarc(g3p,t5b,1).tparc(t5b,dhap,1).
ptarc(g3p,t6,1).tparc(t6,bpg13,2).holds(f16bp;dhap;g3p;bgp13,0,0).
\end{verbatim}
}
we get thousands of answer-sets, for example\footnote{\texttt{\{fires(t1,ts1),\dots,fires(tN,ts1)\}} have been written as \texttt{fires(t1;\dots;tN;ts1)} to save space.}: %
{\small
\begin{verbatim}
holds(bpg13,0,0) holds(dhap,0,0) holds(f16bp,0,0) holds(g3p,0,0) 
holds(bpg13,0,1) holds(dhap,0,1) holds(f16bp,1,1) holds(g3p,0,1) 
holds(bpg13,0,2) holds(dhap,1,2) holds(f16bp,1,2) holds(g3p,1,2) 
holds(bpg13,0,3) holds(dhap,2,3) holds(f16bp,1,3) holds(g3p,2,3) 
holds(bpg13,2,4) holds(dhap,3,4) holds(f16bp,1,4) holds(g3p,2,4) 
holds(bpg13,4,5) holds(dhap,4,5) holds(f16bp,1,5) holds(g3p,2,5) 
fires(t3,0) fires(t3;t4,1) fires(t3;t4;t5a;t5b,2) 
fires(t3;t4;t5a;t5b;t6,3) fires(t3;t4;t5a;t5b;t6,4) 
fires(t3;t4;t5a;t5b;t6,5)
\end{verbatim}
}

\section{Changing Firing Semantics}\label{sec:enc_max}
The ASP code above implements the \textit{set firing} semantics. It can produce a large number of answer-sets, since any subset of a firing set will also be fired as a firing set. For our biological system modeling, it is often beneficial to simulate only the maximum activity at any given time-step. We accomplish this by defining the \textit{maximal firing set} semantics, which requires that a maximal subset of non-conflicting transitions fires at a single time step\footnote{Such a semantics reduces the reachable markings. See \cite{Burkhard1980} for the analysis of its computational power.}. Our semantics is different from the firing multiplier approach used by \cite{PetriNetMaxParallelism}, in which a transition can fire as many times as allowed by the tokens available in its source places. Their approach requires an exponential time firing algorithm in the number of transitions. Our maximal firing set semantics is implemented by adding the following rules to the encoding in Section~\ref{sec:enc_basic}:

{\small
\begin{description}
\item[a5:] \texttt{\small could\_not\_have(T,TS) :- enabled(T,TS), not fires(T,TS), \\
   ptarc(S,T,Q), holds(S,QQ,TS), tot\_decr(S,QQQ,TS), Q > QQ - QQQ.}
\item[a6:] \texttt{\small:-not could\_not\_have(T,TS), enabled(T,TS),
   not fires(T,TS), \\trans(T), time(TS).}
\end{description}
}

Rule $a5$ encodes \texttt{\small could\_not\_have(T,TS)} which means that an enabled transition $T$ that did not fire at time $TS$, could not have fired because its firing would have resulted in overconsumption. Rule $a6$ eliminates any answer-sets in which an enabled transition did not fire, that could not have caused overconsumption. Intuitively, these two rules guarantee that the only reason for an enabled transition to not fire is conflict avoidance (due to overconsumption). With this firing semantics, the number of answer-sets produced for Petri Net in Figure~\ref{fig:q1:a} reduces to 2.

\begin{proposition}
There is 1-1 correspondence between the answer sets of $\Pi^1(PN,M_0,k)$ and the execution sequences of $PN$.
\end{proposition}

Other firing semantics can be encoded with similar ease\footnote{For example, if \textit{interleaved} firing semantics is desired, replace rules $a5,a6$ with the following:
\begin{description}
\item[a5':] \texttt{more\_than\_one\_fires :- fires(T1,TS), fires(T2, TS), T1 != T2, time(TS).}
\item[a6':] \texttt{:- more\_than\_one\_fires.}
\end{description}
}. We now look at Petri Net extensions and show how they can be easily encoded in ASP.

\section{Extension - Reset Arcs}\label{sec:enc_reset}

\begin{definition}[Reset Arc]
A Reset Arc in a Petri Net $PN^R$ is an arc from place $p$ to transition $t$ that consumes all tokens from its input place $p$ on firing of $t$. A Reset Petri Net is a tuple $PN^R = (P,T,E,W,R)$ where, $P, T, E, W$ are the same as for PN; and $R: T \rightarrow 2^P$ defines reset arcs
\end{definition}

\begin{figure}[htbp]
\centering
\vspace{-30pt}
\includegraphics[width=9cm]{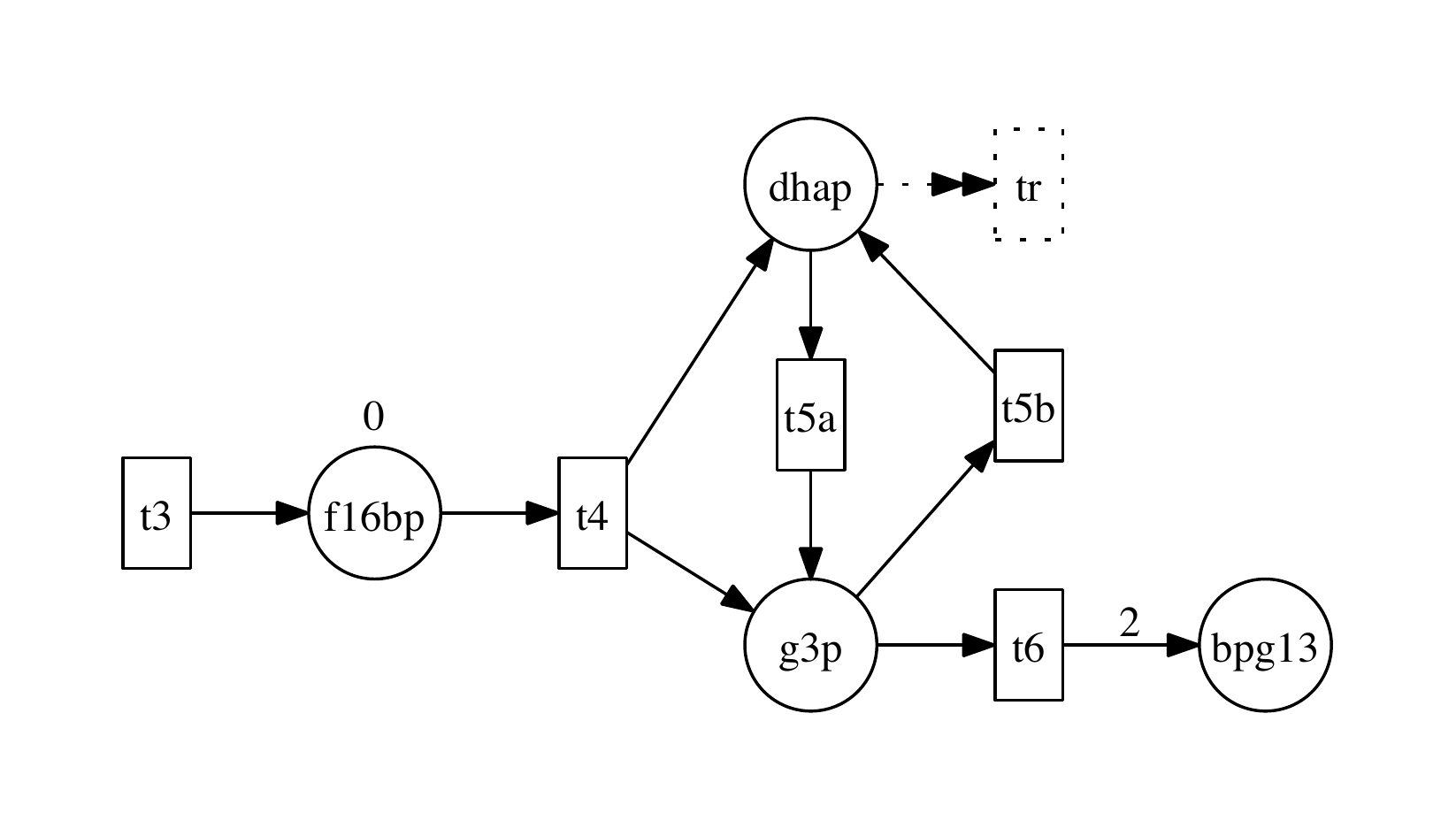}
\caption{Petri Net of Fig~\ref{fig:q1:a} extended with a reset arc from $dhap$ to $tr$ shown with double arrowhead.}
\label{fig:q1:b}
\end{figure}

Figure~\ref{fig:q1:b} shows an extended version of the Petri Net in Figure~\ref{fig:q1:a} with a reset arc from $dhap$ to $tr$ (shown with double arrowhead). In biological context it models the removal of all quantity of compound $dhap$. Petri Net execution semantics with reset arcs is modified for conflict detection and execution as follows:

\begin{definition}[Conflicting Transitions in $PN^R$]
A set of enabled transitions conflict in $PN^R$ w.r.t. $M_k$ if firing them simultaneously will consume more tokens than are available at any one of their common input-places. $T_e = \{ t \in T : enabled_{M_k}(t) \}$ conflict if:
\[
\exists p \in P : M_k(p) < (\displaystyle\sum_{t \in T_e \wedge (p,t) \in E^-}{W(p,t)} + \displaystyle\sum_{t \in T_e \wedge p \in R(t)}{M_k(p)})
\]
\end{definition}

\begin{definition}[Execution in $PN^R$]
Execution of a transition set $T_i$ in $PN^R$ has the following effect:
\[
\forall p \in P \setminus R(T_i), M_{k+1}(p) = M_k(p) -  \sum_{\substack{t \in T_i \wedge p \in \bullet t}} W(p,t) + \sum_{\substack{t \in T_i \wedge p \in t \bullet}} W(t,p)
\]
\vspace{-10pt}
\[
\forall p \in R(T_i), M_{k+1}(p) = \sum_{t \in T_i \wedge p \in t \bullet} W(t,p)
\]
where $R(T_i)=\displaystyle\bigcup_{\substack{t \in T_i}} R(t)$ and represents the places emptied by $T_i$ due to reset arcs.
\end{definition}

Since a reset arc from $p$ to $t$, $p \in R(t)$ consumes current marking dependent tokens, we extend \texttt{\small ptarc} to include time and replace $f3,f4,e1,r1,r2$ with $f6,f7,e3,r6,r7$, respectively in the Section~\ref{sec:enc_max} encoding and add rule $f8$ for each reset arc:
{\small
\begin{description}
\item[f6:] Rules \texttt{\small ptarc($p_i,t_j,W(p_i,t_j),ts_k$):-time($ts_k$).} for each non-reset arc $(p_i,t_j) \in E^-$
\item[f7:] Rules \texttt{\small tparc($t_i,p_j,W(t_i,p_j),ts_k$):-time($ts_k$).} for each non-reset arc $(t_i,p_j) \in E^+$
\item[e3:] \texttt{\small notenabled(T,TS) :- ptarc(P,T,N,TS), holds(P,Q,TS), Q < N,\\
   place(P), trans(T), time(TS), num(N), num(Q).}
\item[r6:] \texttt{\small add(P,Q,T,TS) :- fires(T,TS), tparc(T,P,Q,TS), time(TS).}
\item[r7:] \texttt{\small del(P,Q,T,TS) :- fires(T,TS), ptarc(P,T,Q,TS), time(TS).}
\item[f8:] Rules \texttt{\small ptarc($p_i,t_j,X,ts_k$) :- holds($p_i,X,ts_k$), num($X$), $X>0$.} for each reset arc between $p_i$ and $t_j$ using $X=M_k(p_i)$ as arc-weight at time step $ts_k$.
\end{description}
}

Rule $f8$ encodes place-transition arc with marking dependent weight to capture the notion of a reset arc. 
The execution semantics of our definition are slightly different from the standard definition in ~\cite{araki1976some}, even though both capture similar operations. Our implementation considers token consumption by reset arc in contention with other token consuming arcs from the same place, while the standard definition considers token consumption as a side effect, not in contention with other arcs. 

\begin{figure}
\centering
\vspace{-30pt}
\includegraphics[width=6cm]{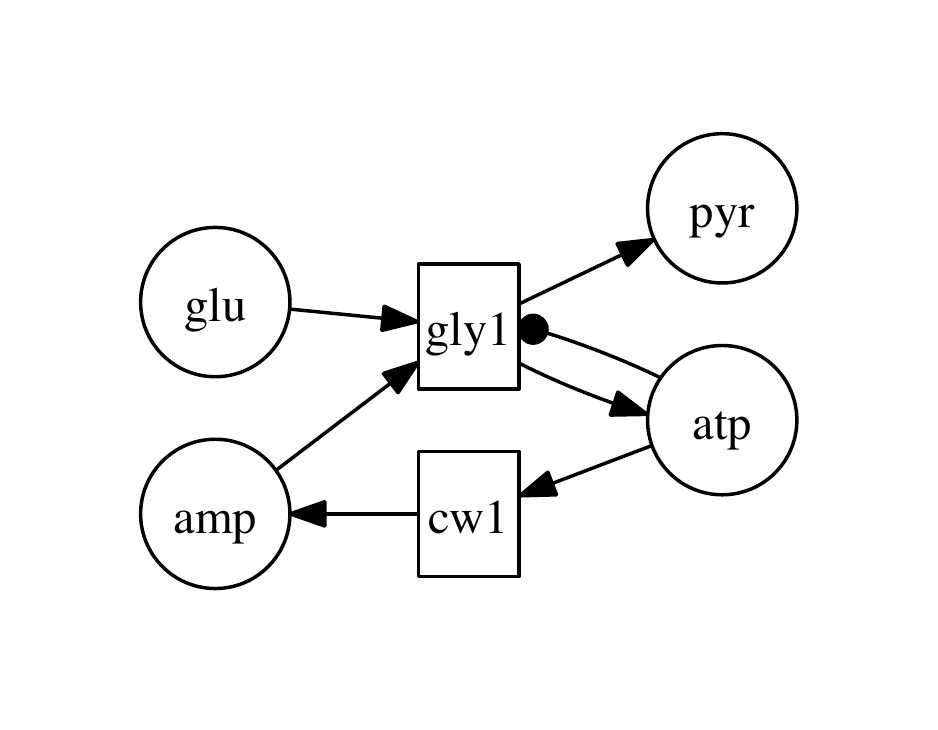}
\caption{Petri Net showing feedback inhibition arc from $atp$ to $gly1$ with a bullet arrowhead. Inhibitor arc weight is assumed $1$ when not specified.} 
\label{fig:q2}
\end{figure}

We chose our definition to allow modeling of biological process that removes all available quantity of a substance in a maximal firing set. Consider Figure~\ref{fig:q1:b}, if $dhap$ has $1$ or more tokens, our semantics would only permit either $t5a$ or $tr$ to fire in a single time-step, while the standard semantics can allow both $t5a$ and $tr$ to fire simultaneously, such that the reset arc removes left over tokens after $(dhap,t5a)$ consumes one token. 

We could have, instead, extended our encoding to include self-modifying nets~\cite{SelfModNets}, but our definition provides a simpler solution. Standard semantics, however, can be easily encoded\footnote{
Standard reset arc can be added to Section~\ref{sec:enc_max} encoding by splitting $r5$ into $r5a',r5b'$ and adding $f8',a7'$ as follows:
\begin{description}
\item[f8':] \texttt{rptarc($p_i$,$t_j$).} - fact capturing a reset arc
\item[a7':] \texttt{reset(P,TS) :- rptarc(P,T), place(P), trans(T), fires(T,TS), time(TS).} - rule to capture if $P$ will be reset at time $TS$ due to firing of transition $T$ that has a reset arc on it from $P$ to $T$.
\item[r5a':] \texttt{holds(P,Q,TS+1) :- holds(P,Q1,TS), tot_incr(P,Q2,TS), tot_decr(P,Q3,TS), Q=Q1+Q2-Q3, place(P),  num(Q;Q1;Q2;Q3), time(TS), time(TS+1), not reset(P,TS).} - rule to compute marking at $TS+1$ when $P$ is not being reset.
\item[r5b':] \texttt{holds(P,Q,TS+1) :- tot_incr(P,Q,TS), place(P),  num(Q), time(TS), time(TS+1), reset(P,TS).} - rule to compute marking at $TS+1$ when $P$ is being reset.
\end{description}
}.

\begin{proposition}
There is 1-1 correspondence between the answer sets of $\Pi^2(PN^R,M_0,k)$ and the execution sequences of $PN^R$.
\end{proposition}

\section{Extension - Inhibitor Arcs}\label{sec:enc_inhibit}
\begin{definition}[Inhibitor Arc]
An inhibitor arc~\cite{PetersonPetriNets} is a place--transition arc that inhibits its transition from firing as long as the place has any tokens in it. An inhibitor arc does not consume any tokens from its input place. A Petri Net with reset and inhibitor arcs is a tuple $PN^{RI}=(P,T,E,W,R,I)$, where, $P, T, E, W, R$ are the same as for $PN^R$; and $I: T \rightarrow 2^P$ defines inhibitor arcs.
\end{definition}

\begin{figure}[htbp]
\centering
\vspace{-25pt}
\includegraphics[width=6cm]{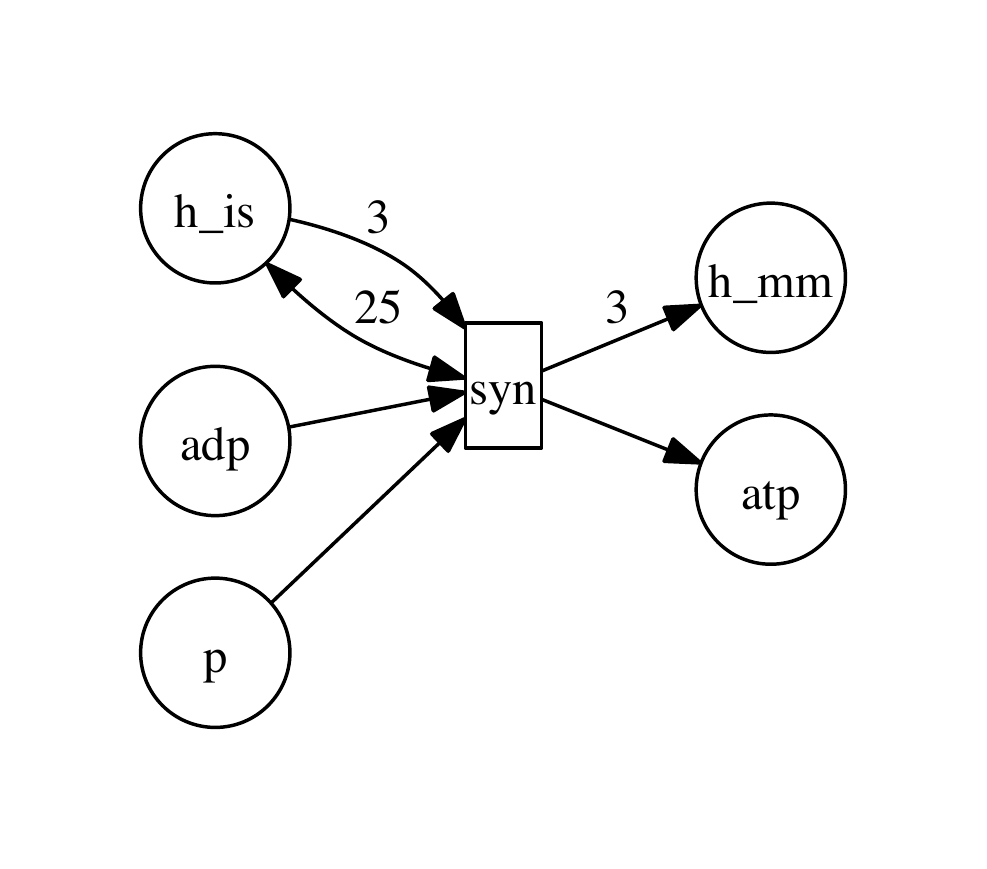}
\caption{Petri Net with read arc from $h\_is$ to $syn$ shown with arrowhead on both ends. The transition $syn$ will not fire unless there are at least $25$ tokens in $h\_is$, but when it executes, it only consumes $3$ tokens.}
\label{fig:atp_synth}
\end{figure}

Figure~\ref{fig:q2} shows a Petri Net with inhibition arc from $atp$ to $gly1$ with a bulleted arrowhead. It models biological feedback regulation in simplistic terms, where excess $atp$ downstream causes the upstream $atp$ production by glycolysis $gly$ to be inhibited until the excess quantity is consumed~\cite{CampbellBook}. Petri Net execution semantics with inhibit arcs is modified for determining enabled transitions as follows:

\begin{definition}[Enabled Transition in $PN^{RI}$]
A transition $t$ is enabled with respect to marking $M$, $enabled_M(t)$, if all its input places $p$ have at least the number of tokens as the arc-weight $W(p,t)$ and all $p \in I(t)$ have zero tokens, i.e. $(\forall p \in \bullet t : W(p,t) \leq M(p)) \wedge (\forall p \in I(t): M(p) = 0)$ 
\end{definition}

We add inhibitor arcs to our encoding in Section~\ref{sec:enc_reset} as follows:

{\small
\begin{description}
\item[f9:] Rules \texttt{\small iptarc($p_i,t_j,1,ts_k$):-time($ts_k$).} for each inhibitor arc between $p_i \in I(t_j)$ and $t_j$.
\item[e4:] \texttt{\small notenabled(T,TS) :- iptarc(P,T,N,TS), holds(P,Q,TS),
   place(P), \\trans(T), time(TS), num(N), num(Q), Q >= N.}
\end{description}
}

The new rule $e4$ encodes another reason for a transition to be disabled (or not enabled). An inhibitor arc from $p$ to $t$ with arc weight $N$ will cause its target transition $t$ to not enable when the number of tokens at its source place $p$ is greater than or equal to $N$, where $N$ is always $1$ per rule $f9$. 

\begin{proposition}
There is 1-1 correspondence between the answer sets of $\Pi^3(PN^{RI},M_0,k)$ and the execution sequences of $PN$.
\end{proposition}

\section{Extension - Read Arcs}\label{sec:enc_query}
\begin{definition}[Read Arc]
A read arc (a test arc or a query arc)~\cite{christensen1993coloured} is an arc from place to transition, which enables its transition only when its source place has at least the number of tokens as its arc weight. It does not consume any tokens from its input place. A Petri Net with reset, inhibitor and read arcs is a tuple $PN^{RIQ}=(P,T,W,R,I,Q,QW)$, where, $P,T,E,W,R,I$ are the same as for $PN^{RI}$; $Q \subseteq P \times T $ defines read arcs; and $QW: Q \rightarrow \mathds{N} \setminus \{0\}$ defines read arc weight.
\end{definition}

Figure~\ref{fig:atp_synth} shows a Petri Net with read arc from $h\_is$ to $syn$ shown with arrowhead on both ends. It models the ATP synthase $syn$ activation requiring a higher concentration of $H+$ ions  $h\_is$ in the intermembrane space~\footnote{This is an oversimplified model of $syn$ (ATP synthase) activation, since the actual model requires an $H+$ concentration differential across membrane.}. The reaction itself consumes a lower quantity of $H+$ ions represented by the regular place-transition arc~\cite{CampbellBook,berg2002proton}. 
Petri Net execution semantics with read arcs is modified for determining enabled transitions as follows:

\begin{definition}[Enabled Transition in $PN^{RIQ}$]
A transition $t$ is enabled with respect to marking $M$, $enabled_M(t)$, if all its input places $p$ have at least the number of tokens as the arc-weight $W(p,t)$, all $p_i \in I(t)$ have zero tokens and all $p_q : (p_q,t) \in Q$ have at least the number of tokens as the arc-weight $W(p,t)$, i.e. $(\forall p \in \bullet t : W(p,t) \leq M(p)) \wedge (\forall p \in I(t): M(p) = 0) \wedge (\forall (p,t) \in Q: M(p) \geq QW(p,t))$
\end{definition}

We add read arcs to our encoding of Section~\ref{sec:enc_inhibit} as follows:
{\small
\begin{description}
\item[f10:] Rules \texttt{\small tptarc($p_i,t_j,QW(p_i,t_j),ts_k$):-time($ts_k$).} for each read arc $(p_i,t_j) \in Q$.
\item[e5:] \texttt{\small notenabled(T,TS):-tptarc(P,T,N,TS),holds(P,Q,TS),\\
    place(P),trans(T), time(TS), num(N), num(Q), Q < N.}
\end{description}
}

The new rule $e5$ encodes another reason for a transition to not be enabled. A read arc from $p$ to $t$ with arc weight $N$ will cause its target transition $t$ to not enable when the number of tokens at its source place $p$ is less than the arc weight $N$.

\begin{proposition}
There is a 1-1 correspondence between the answer sets of $\Pi^4(PN^{RIQ},M_0,k)$ and the execution sequences of $PN^{RIQ}$.
\end{proposition}

\section{Example Use Case of Our Encoding and Additional Reasoning Abilities}

We illustrate the usefulness of our encoding by applying it to the following simulation based reasoning question from \cite{CampbellBook}\footnote{As it appeared in  \texttt{https://sites.google.com/site/2nddeepkrchallenge/}}:

\begin{question}\label{q1}
``At one point in the process of glycolysis, both dihydroxyacetone phosphate (DHAP) and glyceraldehyde 3-phosphate (G3P) are produced. Isomerase catalyzes the reversible conversion between these two isomers. The conversion of DHAP to G3P never reaches equilibrium and G3P is used in the next step of glycolysis. What would happen to the rate of glycolysis if DHAP were removed from the process of glycolysis as quickly as it was produced?''
\end{question}

\begin{figure}
\centering
\vspace{-10pt}
\includegraphics[width=0.75\textwidth]{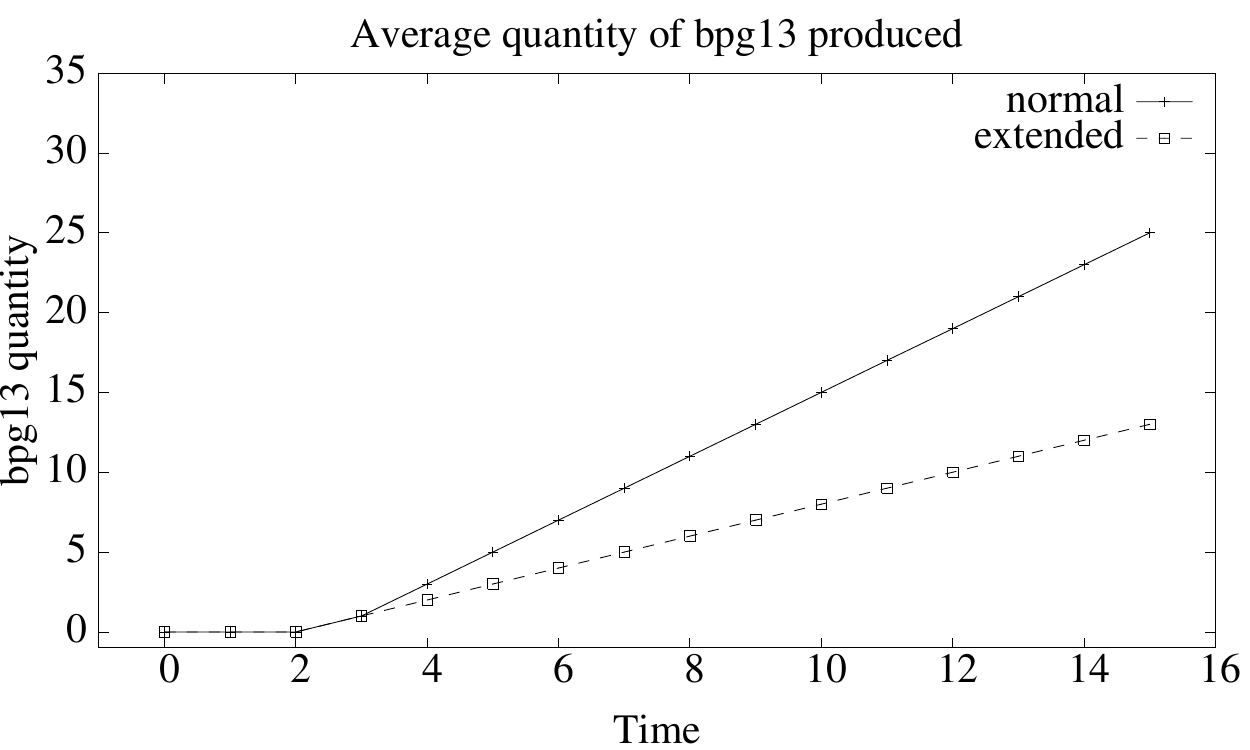}
\caption{Average quantity of $bpg13$ produced by the normal and extended Petri Net models in Fig.~\ref{fig:q1:a} and Fig.~\ref{fig:q1:b} for a simulation length of $15$.}
\label{fig:q1:avg:runs}
\end{figure}

\begin{figure}
\centering
\vspace{-10pt}
\includegraphics[width=0.75\textwidth]{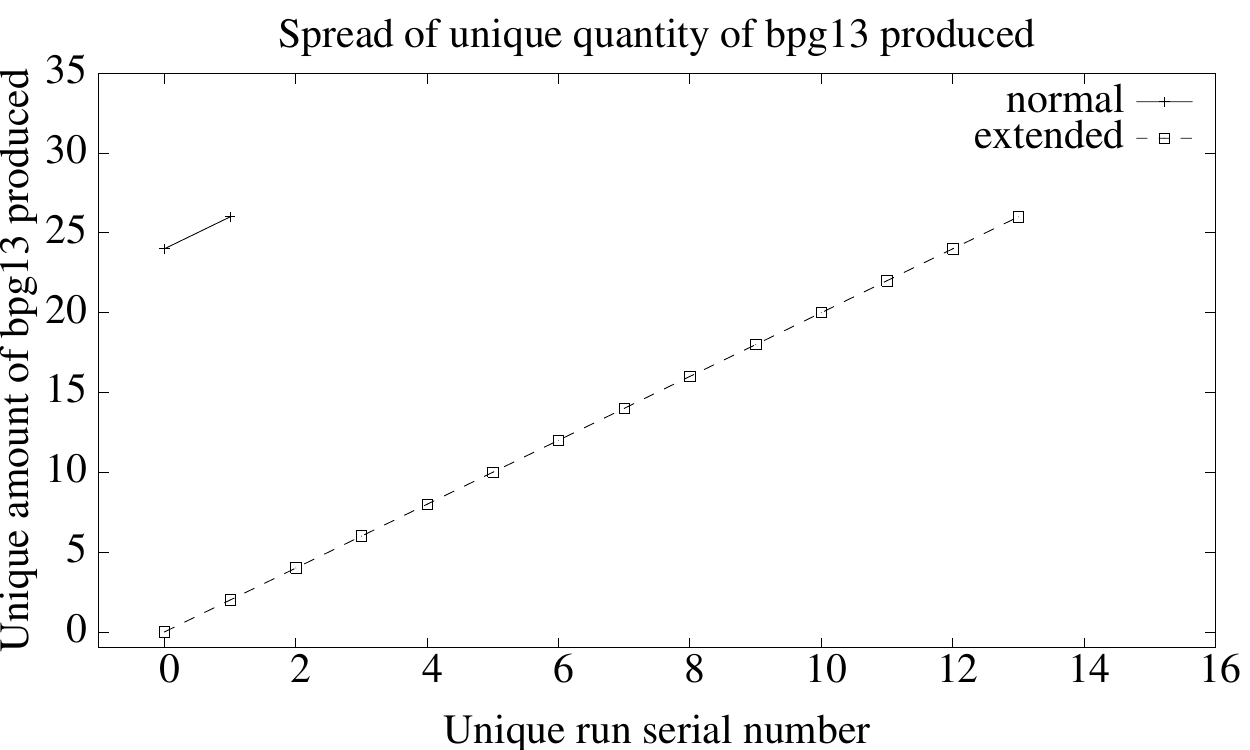}
\caption{Spread of the quantity of $bpg13$ produced by the normal and extended Petri Net models in Fig.~\ref{fig:q1:a} and Fig.~\ref{fig:q1:b} for a simulation length of $15$.}
\label{fig:q1:uniq:spread}
\end{figure}

In order to answer this question, we create a Petri Net model of sub-portion of the normal glycolysis pathway relevant to the question (Figure~\ref{fig:q1:a}) and encode it in ASP using the encoding in Section~\ref{sec:enc_basic}. We then extend this pathway by adding a reset arc to it for modeling the immediate removal of $dhap$ (Figure~\ref{fig:q1:b}) and encode it in ASP using the encoding in Section~\ref{sec:enc_reset}. We simulate both models for the same number of time steps using the maximal firing set semantics from Section~\ref{sec:enc_max}, since we are interested in the maximum change (in $bpg13$ production) between the two scenarios. Figure~\ref{fig:q1:avg:runs} shows the average quantity of $bpg13$ produced by the normal and the extended Petri Net models for a $15$-step simulation and Figure~\ref{fig:q1:uniq:spread} shows the spread of unique $bpg13$ quantities produced during these simulations.  We compute the rate of glycolysis as the ratio ``$bpg13 / ts$'' at the end of the simulation. We post process these results to determine the average rates as well as the spread among the answer-sets. Our results show a lower rate of $bpg13$ production and hence glycolysis in the extended pathway, with a spread of $bpg13$ quantity ranging from zero to the same amount as the normal pathway. Although we are using average rates to answer this question, the ASP encoding produces all possible state evolutions, which may be further analyzed by assigning different probabilities to these state evolutions.

ASP allows us to perform additional reasoning not directly supported by various Petri Net formalisms examined by us~\footnote{We examined a significant number of Petri Net formalisms}. For example, if we are given partial state information, we can use it in ASP as way-points to guide the simulation. Consider that we want to determine the cause of a substance $S$'s quantity recovery after it is depleted. Using ASP, we can enumerate only those answer-sets where a substance $S$ exhausts completely and then recovers by adding constraints. The answer sets produced can then be passed on to an additional ASP reasoning step that determines the general cause leading to the recovery of $S$ after its depletion. This type of analysis is possible, since the answer-sets enumerate the entire simulation evolution and all possible state evolutions.

Although not a focus of this work, various Petri Net properties can be easily analyzed using our ASP encoding.
For example dynamic properties can be analyzed as follows: one can test reachability of a state (or marking) $S$ by adding a constraint that removes all answer-sets that do not contain $S$; boundedness by adding a constraint on token count; basic liveness by extending the Petri Net model with source transitions connected to all places, switching to the interleaved firing semantics, and removing answers where the subject transition to be tested is not fired.
Intuitively, structural properties can be analyzed as follows: T-invariants can be extracted by enumerating transitions $(t_i,\dots,t_{k-1})$ whenever marking $M_k = M_i, k>i$ using an interleaved firing semantics. P-invariants can be extracted by using interleaved firing semantics to visit all possible state (or marking) evolutions, selecting a subset of places $P_i \subseteq P$, and eliminating all answer-sets where $\sum_{p \in P_i}{M_{k+1}(p)} \neq \sum_{p \in P_i}{M_k(p)}, k \geq 0$.

\section{Related Work and Conclusion}
Here we will elaborate a bit on some of the existing Petri Net systems\footnote{The Petri Net Tools Database web-site summarizes a large slice of existing tools \texttt{http://www.informatik.uni-hamburg.de/TGI/PetriNets/tools/quick.html}}, (especially ones used in biological modeling) and put them in context of our research. We follow that with some ways existing Petri Net tools are used for biological analysis and put them in the context of our research. 

CPN Tools~\cite{jensen2007coloured} is a graphical tool for simulating Colored Petri Nets with discrete tokens. It supports guards, timed transitions, and hierarchical transitions but currently does not have direct support for inhibit or reset arcs. It pursues one simulation evolution, breaking transition choice ties randomly. Cell Illustrator~\cite{CellIllustratorBook} is a closed source Java based graphical tool for simulating biological systems using Hybrid Functional Petri Nets (HPFNe). HPFNs combine features from Continuous as well as Discrete Petri Nets. Cell Illustrator only supports uncolored tokens and pursues one simulation evolution, breaking transition choice ties randomly.
Snoopy~\cite{SnoopyPN} is a graphical tool written in C++ for analyzing biological models. It supports simulating Colored, Stochastic, Hybrid, and Continuous Petri Nets with read, reset, and inhibit arcs under a few firing semantics, including maximal firing. However, it does not explore all possible evolutions and it is unclear if simulation results can be exported for further reasoning. Renew~\cite{kummer1999renew} is an open source simulation tool written in Java. It supports Colored tokens, (reference semantics of) Object tokens, guard conditions, inhibit arcs, reset arcs, timed tokens, arc delays, and token reservation. Token nets can be created on the fly. Its use requires some knowledge of Java. Some features like arc pre-delays when combined with transition delays can lead to complicated semantics. It is unclear how ties are broken to resolve transition conflicts and whether all possible state evolutions are explored. 

\cite{Heiner2006,Hofestadt1998,li2006structural} have previously used Petri Nets to analyze biological pathways, but their analysis is mostly limited to dynamic and structural properties. \cite{peleg2005using} surveyed various Petri Net implementations to study the properties and dynamics of biological systems. They defined a series of question that can be answered by Petri Nets using simulation. Contrary to their approach, we picked our questions from college level text books to capture real world questions. Our approach is also different in that we model the questions as Petri Net extensions and we can perform further reasoning on simulation runs.

{\bf Conclusion:} In this paper we discussed the appropriateness of Petri Nets--especially ASP implementation of Petri Nets, for modeling biological pathways and answering realistic questions (the ones found in biology textbooks) with respect to such pathways. We presented how simple Petri Nets can be encoded in ASP and showed that ASP provides an elaboration tolerant way to easily realize various Petri Net extensions and firing semantics. The  extensions include  changing of the firing semantics, and allowing reset arcs, inhibitor arcs and read arcs. Our encoding has a low specification-implementation gap. It allows enumeration of all possible evolutions of a Petri Net simulation as well as the ability to carry out additional reasoning about these simulations. We also presented an example use case of our encoding scheme. Finally, we briefly discussed other Petri Net systems and their use in biological modeling and analysis and compared them with our work. Our focus in this paper has been less on performance and more on ease of encoding, extensibility, exploring all possible state evolutions, and strong reasoning abilities not supported by other Petri Net implementations examined. In a follow on enhanced version of this paper, we will carry out detailed performance analysis. 
In a sequel of this paper, we will present extension of this work to other Petri Net extensions, such as colored tokens, priority transitions, and durative transitions. 

\bibliography{pn_iclp}
\bibliographystyle{unsrt}

\end{document}